\documentclass[10pt,twocolumn,letterpaper]{article}

\usepackage{iccv}
\usepackage{times}
\usepackage{epsfig}
\usepackage{graphicx}
\usepackage{amsmath}
\usepackage{amssymb}

\usepackage[breaklinks=true,bookmarks=false]{hyperref}

\iccvfinalcopy 

\def\iccvPaperID{*****} 

\ificcvfinal\fi
\newcommand{\norm}[1]{\left\lVert#1\right\rVert}
\usepackage{pifont}
\newcommand{\cmark}{\ding{51}}%
\newcommand{\xmark}{\ding{55}}%
\usepackage{multicol}
\begin{document}

\title{View-to-Label: Multi-View Consistency for Self-Supervised 3D Object Detection}

\author{Issa Mouawad*\\
University of Genoa\\
Genoa, Italy\\
{\tt\small issa.mouawad@dibris.unige.it}
\and
Nikolas Brasch\\
Technical University of Munich\\
Munich, Germany\\{\tt\small nikolas.brasch@tum.de}
\and
Fabian Manhardt\\
Google\\
Zurich, Switzerland\\{\tt\small fabianmanhardt@google.com}
\and
Federico Tombari\\
Google\\
Zurich, Switzerland\\{\tt\small tombari@in.tum.de}
\and
Francesca Odone\\
University of Genoa\\
Genoa, Italy\\{\tt\small francesca.odone@unige.it}
}

\maketitle
\ificcvfinal\thispagestyle{empty}\fi

\begin{abstract}
   For autonomous vehicles, 
driving safely 
is highly dependent on the capability to correctly perceive the environment in 3D space, hence the task of 3D object detection represents a fundamental aspect of perception. While 3D sensors deliver accurate metric perception, monocular approaches enjoy cost and availability advantages that are valuable in a wide range of applications.
Unfortunately, training monocular methods requires a vast amount of annotated data. 
Interestingly, self-supervised approaches have recently been successfully applied to ease the training process and unlock access to widely available unlabelled data. While related research leverages different priors including LIDAR scans and stereo images, such priors again limit usability. 
Therefore, in this work, we propose a novel approach to self-supervise 3D object detection purely from RGB sequences alone, leveraging multi-view constraints and weak labels. 
Our experiments on KITTI 3D  dataset demonstrate performance on par with state-of-the-art self-supervised methods using LIDAR scans or stereo images.
\end{abstract}

\section{Introduction}
\label{sec:intro}
Perception pipelines for intelligent agent navigation require a full understanding of the 3D structure of the environment and the surrounding obstacles.
Thereby, 3D object detection is a core problem to solve as it is crucial for autonomous driving to ensure safe driving.
Thanks to recent advances, current models are able to achieve high accuracy in 3D object detection even in very challenging settings.
While early works leveraged 3D active and passive sensors to accomplish this task~\cite{stereo,LIDAR1,LIDAR2,LIDAR3}, monocular approaches are starting to achieve competitive performance~\cite{smoke,monoflex}.
One of the main pillars of this success is large-scale annotated datasets. Monocular approaches benefit exceptionally from data volume and accurate supervision to learn the ill-posed mapping from an RGB image to 3D bounding boxes~\cite{smoke,monoflex,mono3d,monodis}. 
\begin{figure}
\def\svgwidth{\textwidth}
\center
      \includegraphics[width=0.475\textwidth]{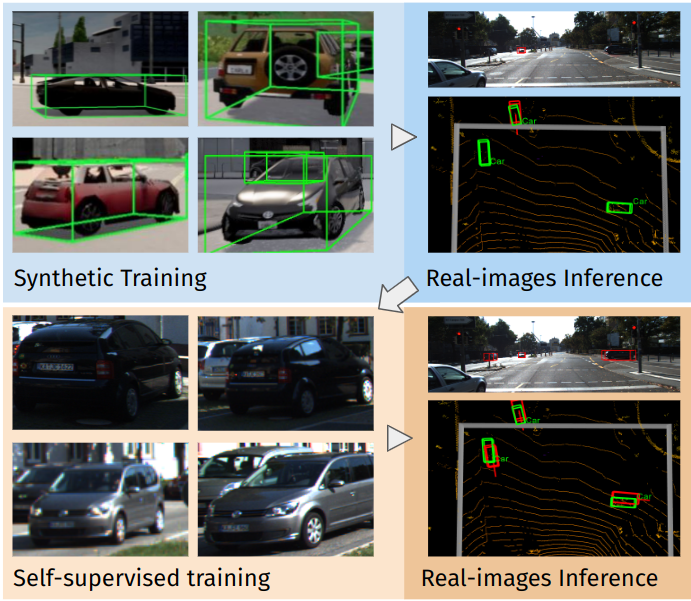}
      
      \caption{Overview of the proposed method. Starting from a pre-training on synthetic data, we fine-tune the model on RGB sequences with no labels obtaining significant improvements in detections (in red) compared to the ground truth (green).}
      \label{fig:mv_losses}
  \end{figure}
Acquiring high-quality labeled data, however, remains a significant bottleneck due to the costs and intense labor needed to annotate and verify 3D labels. This is particularly true when attempting to learn from RGB data alone, as 3D sensors (like LIDAR) are often still necessary to label the data. 
Recently, some works have explored self-supervision for 3D object detection, exploiting different domain priors.
For instance, 3D shapes of certain object classes (e.g. cars) are represented in a low-dimensional space \cite{self6d,monodr,monogrnet}. Raw LIDAR scans, instead, can provide strong geometric cues during training \cite{autolabeling,time-to-label}.

In this work, we aim to train a monocular 3D object detector from RGB sequences alone, leveraging multi-view constraints, in addition to priors on objects' shape and motion patterns. 
In particular, we train a state-of-the-art monocular 3D object detector (\textit{i.e.}~MonoFlex \cite{monoflex}) on synthetic data. Then, we utilize the obtained model to generate pseudo labels on real data. These pseudo labels are subsequently refined using assumptions on objects' motion, relative multi-view constraints, and weak labels generated from pre-trained models. In the core, we propose a differentiable motion-aware warping module to overcome the common rigidity assumption. Furthermore, we make use of multi-view stereo constraints for better 3D perception and we enforce several loss terms in RGB for neighboring views, together with additional terms in the current frames guided by weak labels, such as object mask~\cite{mask} and self-supervised depth estimates~\cite{packnet}, obtained from pre-trained networks.
Eventually, we fine-tune the original model on the obtained pseudo-labels, in an effort to close the gap between the synthetic and real domain (sim-to-real gap). 



We quantitatively evaluate our approach on the common KITTI 3D object detection dataset, utilizing both the train split (the annotated portion of the dataset) and the raw sequences (unannotated) during training. Despite not using any labeled data, our method can achieve results that are close to state-of-the-art self-supervised methods that require LIDAR sensors, and can even rival methods that have been trained fully supervised. 
To summarize, we propose the following contributions,
\begin{itemize}
    \item A novel method to conduct self-supervised 3D object detection from monocular RGB sequences alone, removing the need for labeled data or additional sensors.
    \item Sound multi-view geometry constraints in the context of 3D object detection
    \item Competitive quantitative results on KITTI dataset, and promising qualitative results on DDAD unannotated \cite{packnet} dataset.
    \item A thorough analysis of the prior on scene depth, suggesting future work directions.
\end{itemize}
\section{Related Work}
\subsection{Monocular 3D Object Detection}

According to a recent taxonomy \cite{3dsurvey}, monocular 3D object detection methods can be grouped into methods that lift 2D detections to 3D (referred to as result lifting-based methods) and methods that process 3D features derived from the 2D image plane.
Lifting methods first estimate 2D properties such as the 2D center, the 3D dimensions, and the 3D orientation. The 3D translation is then retrieved by estimating the object's depth, which is then used to back-project the 2D center. 
Commonly. these detectors are built on top of standard region-proposal 2D detectors \cite{faster}, using the RPN features to estimate 3D properties of objects \cite{mono3d,3dop,roi10d,mousavian20173d}.
In contrast, single-stage 2D detectors have been exploited to support efficient 3D inference \cite{monoflex,monodis,m3drpn,movi3d,rtm3d}. Other non-lifting methods, instead, derive features directly in a 3D reference system in the form of point clouds \cite{pseudoLIDAR,pseudoLIDAR++} or projections onto a bird's eye view \cite{philion2020lift,bevdet}.



\subsection{3D Shape Learning in 3D Object Detection}
A core component of almost any self-supervised 3D task is the employed 3D shape prior to represent the objects in the scene.
Thereby, class-level shape priors provide a particularly strong cue for a variety of 3D tasks such as 3D stereo reconstruction \cite{dense_recon} and 3D object detection \cite{pcashapes}. While earlier works use deformable anchor points to represent shapes variety, more recent approaches instead rely on implicit representations, such as truncated signed distance functions (SDF)~\cite{tsdf}. Further, to ease shape estimation, most methods initially learn a low-dimensional continuous shape space, using either variational auto-encoders (VAE) \cite{han2017high,roi10d} or standard PCA~\cite{directshape,pcashapes,dense_recon}.

\subsection{Differentiable and Neural Rendering}
To optimize 3D parameters using 2D images, rendering is oftentimes a core component to directly compare observations with the current parameters. Unfortunately, rendering is commonly not differentiable due to the hard aggregation step during rasterization. Nevertheless, several works have been recently proposed to re-establish the gradient flow during rendering \cite{drsurvey}. 
While one line of work proposes to approximate the gradients~\cite{opendr,gradients2}, another branch of work instead attempts to turn the hard aggregation function into a soft variant~\cite{soft_dr1,soft2,dibrenderer}. In this work, we use the differentiable renderer by~\cite{self6d}, which is based on the Dib-R~\cite{dibrenderer}, combining approximated gradients for background with soft gradients for the foreground, as it provides state-of-the-art results for rendering and additionally allows to also compute depth maps.

\subsection{Self-Supervised 3D Object Detection}
In 3D object detection, self-supervision typically requires a strong prior on the scene, which is usually encoded in the form of initial predictions generated by a pre-trained  detection model \cite{self6d}, or by random initial guesses \cite{monodr}. Exploiting differentiable rendering pipelines \cite{drsurvey}, and render-and-compare losses \cite{3drcnn}, different consistency losses are proposed both in 3D \cite{monodr,autolabeling,koestler} and in 2D using mask, appearance and other photometric cues \cite{koestler,monodr,self6d}. 
Interestingly, recent works either use a combination of RGB images and LIDAR point cloud during training \cite{autolabeling,time-to-label}, or rely solely on LIDAR point clouds to generate 3D pseudo-labels for the task of 3D object detection \cite{lpcg,st3d,caine2021pseudo}.
Despite achieving remarkable results without requiring actual labels for the 3D pose, these approaches heavily rely on LIDAR scans during training, making them rather impractical for real application. Therefore, in this work, we propose to replace LIDAR scans with temporal priors and weak labels established on RGB sequences.

\begin{figure*}
\def\svgwidth{\textwidth}
\center     \includegraphics[width=0.92\textwidth]{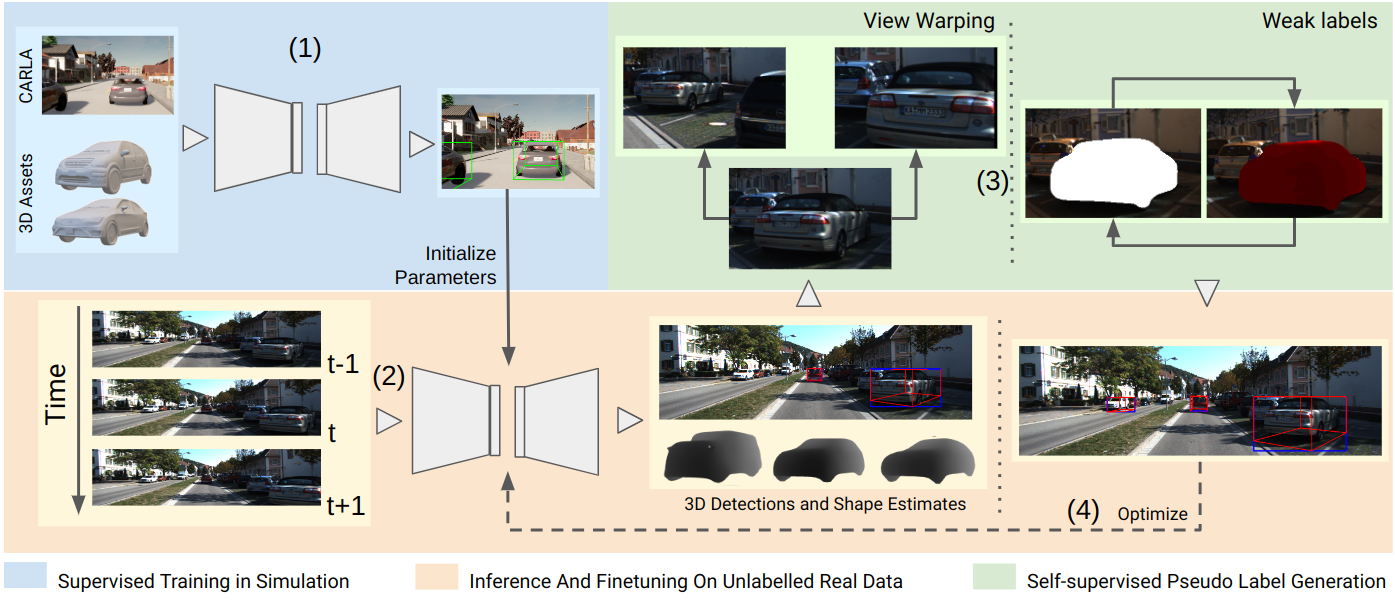}
  \caption{Overview of the overall pipeline. (1) The model is pre-trained on synthetic images generated in Carla. (2) Inference on real images from KITTI provides initial poses. (3) Pseudo-labeling is performed leveraging multi-view consistency and weak labels. (4) The refined poses are used to fine-tune the model}
  \label{fig:mv_method}
\end{figure*}

\section{Methodology}
In this section, we introduce our novel monocular 3D object detector, which can be trained without the use of any ground truth labels for real data and is not dependent on any stereo pair or LIDAR data for self-supervision. Starting from an object's noisy pose defined in frame $i$, we use per-frame weak labels together with different forms of consistencies (including photometric, and mask) towards adjacent views $k$. Intuitively, when warping an object with pose $(t_i,yaw_i)$ from view $i$ to a nearby view $k$, accounting for camera and object motion, one should obtain the respective pose in frame $k$, assuming correct pose estimates.

An abstract overview of our proposed method is provided in Figure~\ref{fig:mv_method}. We first start by pre-training our model in simulation to jointly estimate the object pose and shape (Section \ref{sec:pretraining}). Afterward, we employ our pre-trained model to conduct 3D object detection on real data (Section \ref{sec:inference-on-real-images}). 
We then refine the initial noisy estimates and derive high-quality pseudo-labels (Section \ref{sec:pseudo-labeling}). This is done by relying on the consistency across different views (frames) and using frame-wise weak labels. Eventually, we leverage our obtained pseudo-labels to fine-tune the model in order to bridge the sim-to-real domain gap.

\subsection{Pre-training in Simulation}
\label{sec:pretraining}
As our base monocular 3D object detector, we adopt the MonoFlex model~\cite{monoflex}.
We further follow current related work~\cite{time-to-label} and train our model on synthetic data as generated with the Carla simulator~\cite{carla}. 
Thereby, we train the model fully supervised for 3D object detection and for shape prediction~\cite{pcashapes}, similar to \cite{time-to-label}.
In particular, our model predicts the 2D bounding box $\boldsymbol{b}$, object pose $(\boldsymbol{t},yaw,\boldsymbol{size})$, and shape embedding $\boldsymbol{e}$.

\subsection{Inference on Real Images}
\label{sec:inference-on-real-images}
Upon having finished training in simulation, we leverage the obtained model to run inference on real data in an effort to acquire noisy estimates of poses given the current camera frame $(\boldsymbol{t}_c,yaw_c)$, metric size, and shape embedding $\boldsymbol{e}$.
Note that to better guide the optimization, we leverage additional cues obtained from different pre-trained networks. Specifically, we compute, for latter weak supervision, object masks using a COCO pre-trained Mask R-CNN model~\cite{mask} and object depth maps as estimated by the self-supervised approach PackNet~\cite{packnet}.

Eventually, we construct object trajectories given a video sequence using a simple 3D tracking approach~\cite{ab3dmot}. Subsequently, we classify the objects of each trajectory as either \emph{static} and \emph{moving}, following~\cite{time-to-label}. This is achieved by establishing a world reference system to represent object poses across time. Note that the world-reference system can be relative to any fixed scene point. In this work, we define the camera position and time $0$ as our origin.
\subsection{Self-supervised Pseudo-label Generation}
\label{sec:pseudo-labeling}
In the core, we propose a pseudo-labeling approach that takes as input the detector noisy estimates, mask, and depth cues together with the object trajectories prior, and derives a refined object pose.
The approach employs several loss terms defined either on a single view or on multiple views (Figure \ref{fig:mv_losses}). We first discuss the process of choosing the optimization views and the rendering approach, then we describe in detail each of the loss terms used. 
\begin{figure}
\def\svgwidth{\textwidth}
\center
      \includegraphics[width=0.482\textwidth]{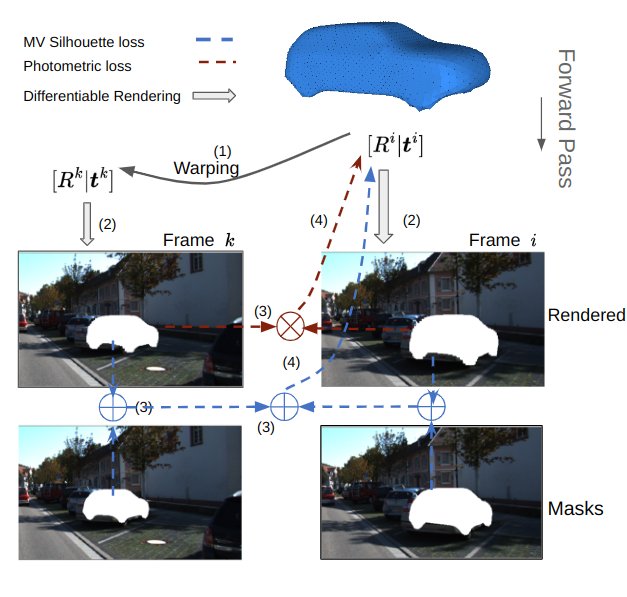}
      
      \caption{Overview of the multi-view optimization: (1) Pose [$R|\boldsymbol{t}$] is warped from frame $i$ to target frame $k$.(2) Retrieved mesh and respective poses are differentiable rendered in each frame. (3) Various losses are calculated using RGB values and object masks. (4) Losses are back-propagated to the pose at frame $i$. Note that we only show multi-view losses in this figure to avoid clutter. }
      \label{fig:mv_losses}
  \end{figure}
\subsubsection{Differentiable Rendering }
\label{sec:dif_render}
To project the pose hypothesis $(\boldsymbol{t}, yaw, \boldsymbol{size})$, along with shape embedding $\boldsymbol{e}$, from the 3D space to the image plane, we require a rendering pipeline. We thus adopt differentiable rendering to further allow backpropagation through the rendering process in order to be able to optimize the pose and shape variables as defined in the 3D reference system. After recovering the object 3D shape from the embedding ${\boldsymbol{e}}$, we use the renderer from \cite{dib,self6d} to render the object mask and depth.

\subsubsection{Views Sampling}
To enforce multi-view constraints, we are required to possess views that exhibit different viewpoints of the object of interest. In addition, these frames need to be close in time to not break the brightness constancy assumption. 
It is well known that choosing multiple and different viewpoints of the object, is beneficial for the optimization procedure as the problem is mathematically better posed~\cite{autolabeling_mvs}.
For a given object pose at time $i$, we sample 4 views covering different local orientation angles (commonly referred to as allocentric rotation, refer to \cite{roi10d} for more details). As for moving objects, we pick the 4 adjacent views around view $i$ to limit the effect of the object's motion.

\subsubsection{Motion-aware Warping}
\label{sec:warping}
When estimating the pose at frame $i$ employing multi-view constraints, we need to be able to transform both the pose $(\boldsymbol{t_i}, yaw_i)$ and objects points $P_{{3D}_i}$, expressed in the camera reference system at frame $i$, to the camera reference system at frame $k$. We refer to the warping operation as $w_{i,k}$, which we require to be fully differentiable with respect to the pose.
 
As apparent motion is caused by a combination of camera motion and  object motion (in the case of non-static objects), the warping operation needs to account for both.

\begin{figure}
\center
      \includegraphics[height=4cm]{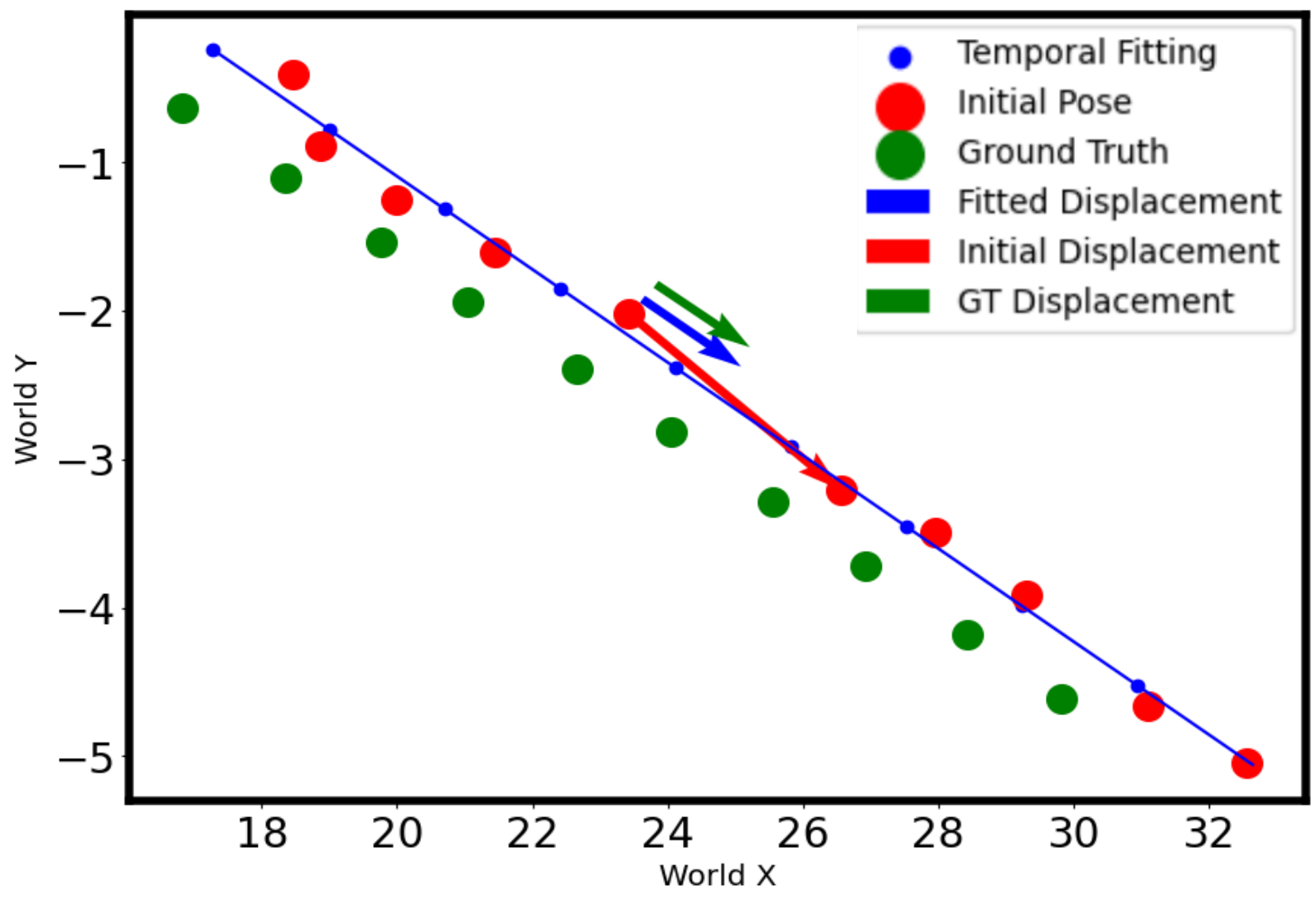}
      
      \caption{ The estimation of the direction and norm of the displacement vector for a moving object. red: the noisy poses and the fitted line. green: the ground truth and a fitted line. blue: the displacement vector placed at an example noisy pose.}
      \label{fig:vel_vector}
  \end{figure}

The camera motion is represented by a rigid transformation, which we obtain directly from the IMU/GPS measurements (but can also be estimated by a structure-from-motion procedure). In KITTI \cite{kitti} dataset, this can be implemented as two transformations: $g_{i,0}$ which transforms 3D points from frame $i$ to frame $0$ (given by the IMU/GPS measurements) and $g_{k,0}^{-1}$; the transformation from frame $k$ to $0$. Formally, we can transform from frame $i$ to $k$ according to
\begin{equation}
\label{equ:itok}
    g_{i,k} = g_{i,0} g_{0,{k}}.
\end{equation}
As for moving objects, warping should also compensate for object motion across frames. This is particularly challenging when initial estimates in the world reference system $\boldsymbol{t_w^i},\boldsymbol{t_w^k}$ are not reliable enough to simply derive a displacement vector between the two positions $\boldsymbol{t}_w^k-\boldsymbol{t}_w^i$.
Instead, we propose to rely on the motion smoothness assumption to acquire a motion direction and a velocity. Specifically, using the piece-wise linearity of the motion, we may recover a vector $\boldsymbol{v}$ expressing the direction of motion which we estimate robustly given a subset of measurements and a RANSAC line-fitting procedure.
The displacement direction (obtained from the line-fitting) is assumed fixed throughout the pose optimization, while the position $\boldsymbol{t}_c^i$ is an optimizable variable.
As for the velocity magnitude (the norm of the displacement vector $\boldsymbol{v}$), we simply assume a local constant velocity, which we calculate by averaging the instantaneous velocities across the portion of the trajectory. Figure \ref{fig:vel_vector} shows that the estimation of the displacement vector is robust, even for noisy pose estimates.

The overall warping transformation is defined as $w_{i,k} = [R|\boldsymbol{t}+\boldsymbol{v}]$ whereas $g_{i,k}=[R|\boldsymbol{t}]$ is the component compensating for the camera motion, while  $\boldsymbol{v}$ is a translational-only component that models object motion (pure translation in a brief time window).
\subsubsection{Optimization Objectives}
We apply an object-centric optimization, treating each trajectory independently. At each time step $i$ of a specific trajectory, we perform a multi-view optimization to refine the respective pose. In particular, we optimize for the translation in the camera frame $\boldsymbol{t}_c^i$, and the metric size, while we keep $yaw_c^i$ fixed using the tracklet smoothness assumption.
Thanks to the differentiable renderer, we are able to derive the losses with respect to the mentioned variables.

\subsubsection*{Silhouette Loss}
The silhouette loss has been used widely to perform 2D alignment between masks, both in pre-deep learning works~\cite{sil_1} as well as in more recent approaches~\cite{monodr}. 

As for the silhouette, we employ object segmentation masks, obtained from Mask RCNN, as weak labels. For each object at frame $i$, we obtain a binary mask $Mask_{fg}^i$ that indicates foreground pixels (pixels belonging to the object), while the rest of the image is considered background $Mask_{bg}^i$. 
As depicted in Figure\ref{fig:mv_method}, and discussed in Section \ref{sec:dif_render}, we render the mask of the predicted object pose and shape $Ren^i_{prob}$, where the value of each pixel of the rendered mask is the probability of the pixel belonging to the mesh surface. 
Eventually, we use the binary cross-entropy loss to supervise the pose via
\begin{equation}
\begin{split}
    \mathcal{L}_{sil} &= - log(Ren_{prob}^i\times Mask_{fg}^i \\
    &+ (1-Ren_{prob}^i)\times Mask_{bg}^i).
    \end{split}
\end{equation}

Since the silhouette loss is only evaluated at frame $i$, the optimization can thus suffer from scale ambiguity, as the projection of the surface depends on both the distance and the size. Nonetheless, this loss is still very useful for the horizontal alignment of the segmentation.

\subsubsection*{Multi-view Silhouette Loss}
\label{mv_sil}
In order to account for scale ambiguity, we propose extending the silhouette loss to the multi-view settings.
Given the differentiable warping transformation $w_{i,k}$, we warp the pose from view $i$ 
$(yaw_c^{i},\boldsymbol{t}_c^{i})$ to view $k$ and obtain $(yaw_c^{i,k},\boldsymbol{t}_c^{i,k})$ as:
\begin{equation}
    (yaw_c^{i,k},\boldsymbol{t}_c^{i,k}) = w_{i,k}(yaw_c^{i},\boldsymbol{t}_c^{i}).
\end{equation}
To simplify the warping notation, we denote the rotation angle around the vertical axis (yaw) as a rotation matrix $R \in SO(3)$. Hence, we represent the warping operation as a matrix multiplication in homogeneous coordinates. 
Subsequently, we render the mask $Ren_{prob}^{i,k}$ and can calculate the multi-view silhouette loss using views $i$ and $k$:
\begin{equation}
\begin{split}
    \mathcal{L}_{mv\_sil} &=
    \sum_k - log(Ren_{prob}^{i,k}\times Mask_{fg}^{k} \\ &+ (1-Ren_{prob}^{i,k})\times Mask_{bg}^{k}).
    \end{split}
\end{equation}

Note that due to our differentiable warping operation, we are still able to propagate the gradients from different views to the original pose at frame $i$.

\subsubsection*{Photometric Loss}
\label{mv_photo}
Aside from the weak supervision provided by object masks, raw RGB values can provide an additional supervision signal, relying on the photometric consistency between nearby frames. At the 3D pose level, pixels belonging to an object should be photometrically consistent across views, assuming a correct 3D pose. In theory, we can identify pixels belonging to the same object, and verify that their RGB values are coherent when warped to a different view.
In practice, it is not very straightforward to warp 2D pixels using transformations defined in a 3D reference frame. 

To this end, starting from a view $i$, and given a pose hypothesis ($\boldsymbol{t}_c^i, yaw_c^i, \boldsymbol{size}^i$) together with the shape embedding $\boldsymbol{e}^i$, we differentiably render depth values $D$ of the 3D mesh, which effectively assigns pixels $P_{2D}$ with their corresponding depth value (according to the pose hypothesis).
The back projection $\pi^{-1}(P_{2D},D)$ can then yield points in the camera reference system, which can be warped using $w_{i,k}$.
The photometric loss can, hence, be formulated using some loss function such as $L_1$ or a smoother version according to

\begin{equation}
    \mathcal{L}_{photo} =\sum_k \norm{P_{2D} - \pi_{k} (w_{i,k}  \pi^{-1}_i(P_{2D},D) ) }.
\end{equation}

\subsubsection*{Depth Loss}
Certain scenarios may lead to an ill-posed multi-view optimization. For example, the assumptions made about the motion of cars might not always be satisfied. Moreover, some objects might be heavily occluded throughout the tracklet, resulting in very weak or no poses. 

Therefore, we propose to additionally ground our optimization with weak 3D labels from monocular depth estimation. To this end, we first filter the depth map using the obtained object mask to retrieve the object depth mask $D^{\prime}$. Next, we back-project these pixels (along with depth values) to obtain the respective 3D points in the camera system $P_{3D} = \pi^{-1}(P_{2D},D^{\prime})$. After filtering out outliers, we calculate the mean point of this cloud $\bar{P}_{3D}$, corresponding to the central point of the object. We then retrieve the object's mesh $M^i$ from the predicted embeddings $\boldsymbol{e}^i$, which is composed of vertices $P^i$ and faces $Faces^i$. We calculate the center of the vertices $\bar{P}^i$ and we seek alignment with the computed depth center following an L2 loss as

\begin{equation}
    \mathcal{L}_{depth} = \norm{\bar{P}_{3D} - \bar{P}_i}_2^2.
\end{equation}

\subsubsection*{Domain Prior Losses}
While the losses above can indeed help improving the pose optimization, we still sometimes end up in local minima. To overcome this limitation, we thus further enforce additional priors from the literature.

\paragraph*{Vertical point loss}
Following the common planar assumption of the road \cite{directshape,mousavian20173d}, we can explicitly penalize the pose for out-of-plane translation using an $L_2$ loss with
\begin{equation}
    \mathcal{L}_y = \norm{y_i - y_{plane}}_2^2.
\end{equation}
\paragraph*{Size regularization}
Further, given the limited intra-class variation in object sizes for cars, we use the statistics (obtained from the target dataset or any other similar one) to regularize the size variable to a mean value. This helps avoid extreme cases of scale ambiguity.
\begin{equation}
    \mathcal{L}_{size} = \norm{\boldsymbol{size}^i - \boldsymbol{size}_{mean}}_2^2.
\end{equation}

To summarize, the self-supervised loss that we use is made up of the following terms
\begin{equation}
\begin{split}
    \mathcal{L} &= \lambda_{sil} \mathcal{L}_{sil} + \lambda_{mv\_sil} \mathcal{L}_{mv\_sil}+ \lambda_{depth} \mathcal{L}_{depth}\\  &+ \lambda_{photo} \mathcal{L}_{photo}+
    \lambda_{size} \mathcal{L}_{size}+
    \lambda_{y} \mathcal{L}_{y}
    \end{split}
\end{equation}
 \section{Experiments}
In this section, we will present our experimental evaluation. To this end, we will first introduce our experimental setup, before showing our quantitative and qualitative results, demonstrating the usefulness of our contributions.

\subsection{Experimental Setup}
\subsubsection{Datasets}
\paragraph*{KITTI 3D dataset} \cite{kitti}
consists of around ~80k images, of which approximately 15k are labeled with both frame-level 3D and 2D detections, and trajectory identities.
The images are accompanied by synchronized point clouds and other onboard kinematics such as IMU sensors beside GPS localization.
Note that only half of the labels (around 7.5k images) are released with the dataset, the rest is part of the hidden test set. 
Furthermore, most works split the training dataset again into a training and a validation split with around 3.7k samples each~\cite{split}, often referred to as trainsplit and valsplit. 

\textbf{KITTI Raw} offers a large number of additional sequences for which no manual labels are available. Although not possible to  be directly used for training or fine-tuning using standard supervised learning, it has been recently often utilized as a part of different semi-/self-supervised training paradigms.
\paragraph*{Dense Depth for Autonomous Driving DDAD}
The DDAD dataset \cite{packnet} was recently proposed with the primary objective of dense depth estimation. DDAD comprises 200 sequences with synchronized LIDAR scans and camera poses. Despite having no 3D box labels, we leverage DDAD dataset to qualitatively evaluate our method and to demonstrate the generalization of our approach. Note that DDAD is, besides KITTI, the only dataset with pre-trained models for self-supervised depth estimation, a core prior that is required for our method. In particular, we again use the released PackNet~\cite{packnet} model to obtain depth priors.

\subsubsection{Metrics}
To measure the performance of 3D object detection, we employ the typically used \textit{average precision AP} metric, calculated across different recall points. Positive predictions are assigned based on an \textit{intersection-over-union IoU} threshold (which we set to $0.5$ as commonly done in monocular approaches). The overlap can either be calculated between the two 3D boxes (i.e.\ \textit{AP3D}) or between their projection on a horizontal plane, also referred to as \textit{AP} bird's-eye-view (\textit{AP bev}).
In KITTI dataset, the objects are further divided into three difficulty levels (i.e.\ \textit{easy}, \textit{moderate},\textit{ hard}) based on the distance and size in the image plane
\subsection{Pseudo Labeling}
As a first step to assess the effectiveness of the proposed optimization procedure, we evaluate the optimized poses directly against the ground truth labels.
Specifically, we run a pseudo-labeling procedure on sequences belonging to the trainsplit, setting loss weights to: $\lambda_{sil} = \lambda_{mv\_sil} = \lambda_y = 1$, $\lambda_{depth} = \lambda_{size} = 0.5$ and $\lambda_{photo} = 10$, which we find to produce balanced loss terms.
Table~\ref{table:mv_pseudo_labels} reports the average precision calculated for the trainsplit (using the ground truth solely for evaluation) where we compare our proposed method with other state-of-the-art approaches that require LIDAR during pseudo-labeling~\cite{time-to-label,autolabeling}. It is clear that our multi-view approach has a clear disadvantage since we rely only on RGB images, whereas \cite{time-to-label} and \cite{autolabeling} possess 3D information from the LIDAR. This can be also observed in the accuracy gap we notice across different pseudo-labeling iterations. Nevertheless, the positive trend demonstrates the capacity of our proposed supervision for the generation of improving pseudo-labels. To the best of our knowledge, there are no other RGB-only approaches, producing pseudo-labeling on the KITTI train split.
\begin{table}
\centering
\small
\begin{tabular}{*8c}
 \hline
   Variant&\multicolumn{3}{c}{AP 2D \%} & \multicolumn{3}{c}{AP BEV \%}\\
 \hline\hline
 
&Easy&Mod&Hard&Easy&Mod&Hard\\ 
\hline
\multicolumn{7}{c}{1st Iteration}\\
\hline
Ours&80.6&60.0&51.5&46.8&31.8&27.7\\
TTL~\cite{time-to-label}&84.5&63.2&56.0&66.7&45.0&37.9\\

\hline
\multicolumn{7}{c}{2nd
Iteration}\\
\hline
Ours&82.2&66.7&58.0&54.8&39.1&35.0\\
TTL~\cite{time-to-label}&91.5&67.3&57.6&87.2&60.5&50.8\\

SDF~\cite{autolabeling}& \multicolumn{3}{c}{Ground truth boxes}&77.8&59.7&N/A\\
\hline\hline
\end{tabular}
\caption{Comparative analysis on the generated pseudo-labels. Both \cite{autolabeling} and \cite{time-to-label} use lidar during pseudo-labeling.}

\label{table:mv_pseudo_labels}
\end{table}
\subsection{Fine-tuning and Validation}
Eventually, we aim to use the pseudo-labels to fine-tune the synthetically trained model and bridge the domain gap without needing any ground truth labels. In this work, we achieve this objective without any access to 3D sensors. We consider this the main contribution of this work, as RGB videos are more pervasive, and acquiring them is far less demanding than deploying LIDAR scanners. 
Concretely, we alternate between pseudo-label generation and fine-tuning the model. The fine-tuned model is again used to generate the initial poses, as illustrated in Figure \ref{fig:mv_method}.
In Table~\ref{table:mv_val_11}, we report the average precision of the best-fine-tuned model (as obtained after the two iterations) on KITTI validation split, and compare it to other supervised and self-supervised approaches. Note that we also indicate for each method if they utilized LIDAR point clouds in any form during training or inference. Despite reporting overall a bit lower performance, our method proves effective in learning from pure RGB input and demonstrates the importance of further investigation of multi-view-based self-supervision. 
In addition, we perform an experiment leveraging the unannotated KITTI Raw sequences (including the trainsplit) to simulate access to a larger dataset. Results (shown in a separate row) demonstrate the boost our approach receives (around $9$ points for easy examples and $7$ for moderate ones) when using a larger amount of data for training, highlighting the efficacy and scalability of self-supervised approaches.

To our knowledge, no other work addresses self-supervised 3D object detection from purely RGB data on KITTI dataset. Other works which solve the problem under similar conditions, such as~\cite{autolabeling_mvs,cosypose}, perform experiments on constrained indoor datasets, for example, LineMOD \cite{linemod} or Occlusion \cite{occlusion_dataset}. When comparing with MonoDR~\cite{monodr}, we report on par or slightly worse results. However, MonoDR fine-tunes the poses for each frame of the validation set individually by means of differentiable rendering, requiring around 3 minutes per image on average, rendering the method impractical for real applications. In contrast, our method can run inference near real-time (over 14fps), upon having completed our proposed self-supervision.

\begin{table*}

\resizebox{0.95\textwidth}{!}{%
\begin{tabular}{l*{10}{c}}
\hline
  
  &&&\multicolumn{3}{c}{ $AP_{BEV}$ / $AP_{3D}$ ($AP_{R11}$@ 0.5 IoU) }&\multicolumn{3}{c}{ $AP_{BEV}$ / $AP_{3D}$ ($AP_{R40}$@ 0.5 IoU) }&Inference\\
  Method&GT?&Images&Easy&Mod&Hard&Easy&Mod&Hard&(seconds)\\
 \hline
 \multicolumn{10}{c}{Lidar + RGB}\\
 \hline

LPCG~\cite{lpcg}&\cmark&Train&\textbf{67.66/61.75}&\textbf{52.27}/\textbf{49.51}&\textbf{46.65}/\textbf{44.70}&-&-&-&0.16\\
SDFLabel~\cite{autolabeling}& \xmark&Train&51.10/32.90&34.50/22.10&-&-&-&-&-\\
TTL~\cite{time-to-label}&\xmark&Train&52.43/36.71&37.55/26.74&31.21/22.09&48.59/32.10&31.45/21.12&24.40/15.92&0.06\\
TTL~\cite{time-to-label}&\xmark&Raw&\underline{63.94/51.90}&\underline{42.29/33.24}&\underline{35.31/30.39}&\underline{59.63}/\underline{46.95}&\underline{38.31}/\underline{30.08}&\underline{30.62}/\underline{24.41}&0.06\\

\hline
\multicolumn{10}{c}{RGB only}\\
\hline
D3DBBox~\cite {mousavian20173d}&\cmark&Train&30.02/27.04&23.77/20.55&18.83/15.88&-&-&-\\
Mono3D~\cite{mono3d} &\cmark&Train&30.50/25.19&22.39/18.20&19.16/15.52&-&-&-&-\\
M3D-RPN ~\cite{m3drpn} &\cmark&Train&55.37/48.96&42.49/39.57&35.29/33.01&53.35/48.53&39.60/35.94&31.76/28.59&0.16\\
MonoFlex~\cite{monoflex}&\cmark&Train&\textbf{68.62}/\textbf{65.33}&\textbf{51.61}/\textbf{49.54}&\textbf{49.73}/\textbf{43.04}&\textbf{67.08}/\textbf{61.66}&\textbf{50.54}/\textbf{46.98}&\textbf{45.78}/\textbf{41.38}&0.06\\

MonoDR~\cite{monodr}&\xmark&*&51.13/\underline{45.76}&\underline{37.29/32.31}&30.20/\underline{26.19}&\underline{48.53/43.37}&\underline{33.90/29.50}&25.85\underline{22.72}&180\\

\textbf{Ours} &\xmark&Train&42.73/36.94&30.43/25.51&27.56/22.37&40.75/32.43&27.12/21.55&22.05/18.22&0.06\\
\textbf{Ours} &\xmark&Raw&\underline{51.40}/37.92&37.15/28.21&\underline{32.14}/23.67&48.20/33.86&32.03/23.21&\underline{26.88}/19.85&0.06\\
\hline
 \end{tabular}

}
\caption{Average precision with 11 recall points ($R_{11}$) and 40 recall points ($R_{40}$) on KITTI validation set of our method and other supervised and unsupervised methods. (*) \textit{Optimized directly on the validation set}}

\label{table:mv_val_11}
\end{table*}

\subsection{DDAD Experiment}
\begin{figure}
\def\svgwidth{\textwidth}
\center
      \includegraphics[width=0.47\textwidth]{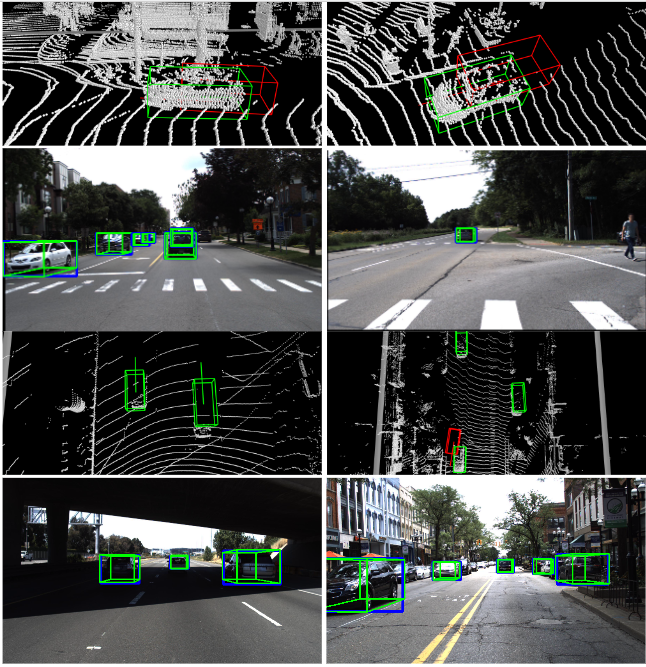}
      
      \caption{Qualitative examples of our fine-tuned model on DDAD \cite{packnet} (green) compared to the original model trained on Carla (red). 2D bounding boxes are shown on the image for  reference (blue)}
      \label{fig:ddad}
  \end{figure}

In order to showcase our pipeline on a different dataset, we utilize the DDAD dataset by \cite{packnet}. To start from the very same pre-trained model, we first scale the image from DDAD dataset \cite{packnet} to the resolution of KITTI and adapt the camera intrinsics. We then generate pseudo-labels for sequences [100,..199], and use them to fine-tune the model on DDAD. As a test set, we use sequences [0,..19]. Given the absence of 3D annotations, we only evaluate qualitatively on DDAD. 
In Figure~\ref{fig:ddad}, we present the results of our method on the test before and after self-supervision. Our results show significant improvements in both 2D detection as well as 3D pose accuracy, we include more examples in the supplementary materials.
 
\section{Scene Depth Analysis}

\subsection{Depth Filtering}

By relying on depth prediction from a self-supervised network, PackNet \cite{packnet}, we inevitably introduce associated errors and noise to the optimization pipeline. Despite performing well on benchmark data, self-supervised monocular depth estimation methods are known to struggle with objects (especially moving in the same direction as the camera) \cite{monodepth1} and their boundaries, making them more vulnerable in our use case. 

To reduce the negative impact of wrong estimates, we pre-process depth values in order to filter out obvious noisy depth values. Specifically, after obtaining the central point depth value, we discard the depth loss if the central point is outside the depth range in KITTI (between 0 and 80 meters distance). We additionally require the central point to be within 6 meters from the initial estimate as those deviate less than per-pixel depth which in some occasions suffer large errors. In the supplementary materials, we include more analysis of the depth errors and their distribution.

\subsection{Semi-Supervised Depth}
In order to verify the positive impact of a higher-accuracy depth, we investigate the use of a semi-supervised depth estimation variant of PackNet presented in \cite{semidepth}. While training the depth network in a joint self-supervised and semi-supervised fashion, the authors leverage highly sparse ground truth depth maps, containing less than 100 points per image accounting for only 0.42\% of valid depth values per image, or 0.06\% of the pixels. 

We perform the pseudo-labeling using the improved depth maps obtained from the semi-supervised network.  Table \ref{tbl:different-depth} demonstrates the capacity of our approach to improve with better depth sources, and the ability to bridge the gap with the LIDAR-based approach. Besides the overall comparison, we also perform a motion-aware evaluation of each of the variants. To split the detected objects into static/moving, we rely on matches between the detections and ground truth objects for which motion classes can be easily assigned. Unmatched detections (false positives) are added to both splits as a worst-case scenario.
Our analysis points to a noticeable decline in accuracy in moving cars (more so in the self-supervised depth case), which can probably be attributed to challenging warping (due to noise in the initial estimates) and frequent failure cases in scene depth in these cases.

\begin{table}
\centering
\small
\begin{tabular}{|c|ccc|ccc|}
 \hline
  Variant&\multicolumn{3}{c|}{AP 3D \% @0.5IoU} & \multicolumn{3}{c|}{AP BEV \% @0.5IoU}\\
 \hline\hline
&Easy&Mod&Hard&Easy&Mod&Hard\\ 
\hline
\multicolumn{7}{|c|}{Static}\\
\hline
self-depth&16.77&15.25&12.54&58.14&37.35&30.99
\\
semi-depth&15.43&13.10&10.83&59.78&40.24&31.51\\

\cite{time-to-label}&15.57&11.99&10.48&74.39&55.95&49.53
\\\hline
\multicolumn{7}{|c|}{Moving}\\
\hline
self-depth&11.20&6.97&7.04&33.45&22.30&22.23
\\
semi-depth&28.45&17.00&16.12&54.65&35.69&33.64\\
\cite{time-to-label}&29.30&18.71&17.31&55.30&40.15&37.83\\

\hline

\multicolumn{7}{|c|}{Overall}\\
\hline
self-depth&15.61&12.17&10.47&46.80&31.80&27.76
\\
semi-depth&25.03&15.25&14.32&59.53&37.87&33.57\\
\cite{time-to-label}&23.94&15.24&13.43&66.50&50.47&46.01\\
\hline

\end{tabular}
\caption{Evaluation of the pseudo-labels by motion class of the three proposed variants of scene geometry: self-supervised depth masks, referred to as self-depth (predicted by \cite{packnet}), semi-supervised depth masks, referred to as semi-depth (predicted by \cite{semidepth}), and raw LIDAR scans, following \cite{time-to-label}}

\label{tbl:different-depth}
\end{table}

  \section{Discussion}
In this work, we have proposed a novel method that can learn 3D object detection from unlabelled RGB sequences, without requiring any LIDAR data or stereo image pairs.
We further showed that leveraging ideas from multi-view geometry proves to be a useful tool for recovering 3D poses of objects. In addition, poses predicted by a model that is initialized in simulation and refined by our optimization pipeline, can produce a strong supervision signal for bridging the sim-to-real domain gap. 
We find that dense depth acquired from self-supervised networks adds useful 3D information (even when noisy) to the optimization, enabling the convergence to better poses, especially for ill-posed situations. 

\clearpage
\twocolumn[{%
 \centering
 \LARGE Supplementary Material} 
 \vspace*{15px}]


\setcounter{page}{1}
\setcounter{section}{0}
\renewcommand{\thesection}{\Alph{section}}

\makeatletter



\section{Scene Depth Ablation}
In this experiment, we disable the self-supervised loss supervision. Specifically, we run the pseudo-labeling procedure for one iteration, discarding the center depth loss, and we use the resulting pseudo-labels to finetune the model. Results, reported in Table \ref{tbl:nodepth}, demonstrate the effectiveness of our approach, and the benefit that depth, even when noisy, brings to the convergence.

\begin{table}[!htp]
\centering
\small
\begin{tabular}{|c|ccc|ccc|}
 \hline
  Variant&\multicolumn{3}{c|}{AP 3D \% @0.5IoU} & \multicolumn{3}{c|}{AP BEV \% @0.5IoU}\\
 \hline\hline
&Easy&Mod&Hard&Easy&Mod&Hard\\ 

w/ Depth&29.45&21.39&19.39&38.15&27.52&23.53
\\
No Depth&15.12&11.23&9.42&19.65&14.59&12.43
\\\hline

\end{tabular}
\caption{Ablation study of the depth loss. We compare the fine-tuned models in both cases on the validation set after the first iteration of pseudo-labeling.}

\label{tbl:nodepth}
\end{table}
\section{Scene Depth Analysis}
\begin{figure}[!htp]
\center
      \includegraphics[height=4cm]{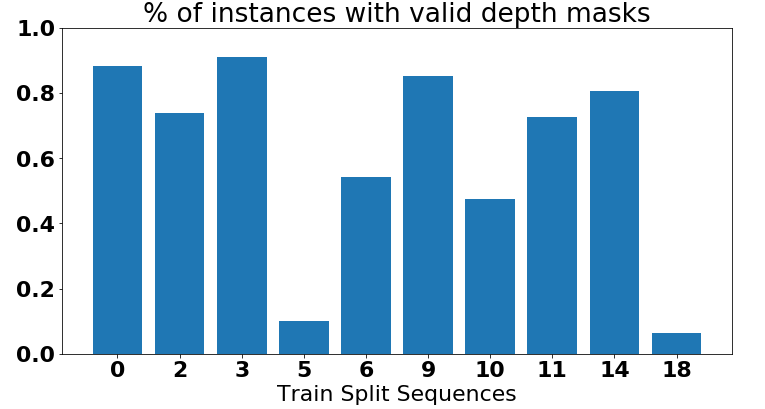}
      
      \caption{The percentage of accepted depth masks of each train split sequence}
      \label{fig:depth-discarded}
  \end{figure}

\begin{figure*}[!htp]
\def\svgwidth{\textwidth}
\center
      \includegraphics[height=5cm]{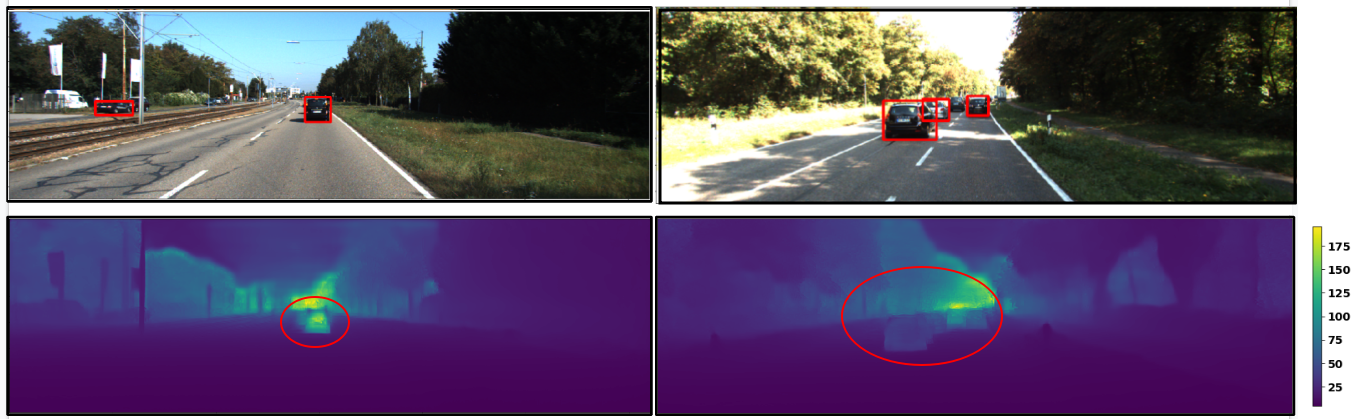}
      
      \caption{Predicted depth failure cases. Top: RGB image, bottom: depth map. The two frames belong the sequences 5 (left) and 18 (right). Both depth maps show incorrect depth prediction around cars in the same lane.}
      \label{fig:vis-depth}
  \end{figure*}

When inspecting the discarded object depth maps (according to the procedure detailed in the main paper Section 5.2), we observe high percentages in two of the train split sequences (5 and 18) as reported in Figure \ref{fig:depth-discarded}.
In Figure \ref{fig:vis-depth}, we illustrate some failure cases of the depth estimation network which predicts incorrect depth for some portions of the scene.
This scenario highlights a typical challenge for self-supervised depth estimation, related to the motion parallax when objects move at a speed close to the one of the camera.

\subsection{Self-Supervised Depth Accuracy}
\label{sec:self}
\begin{figure}[!htp]
\center
      \includegraphics[height=4cm]{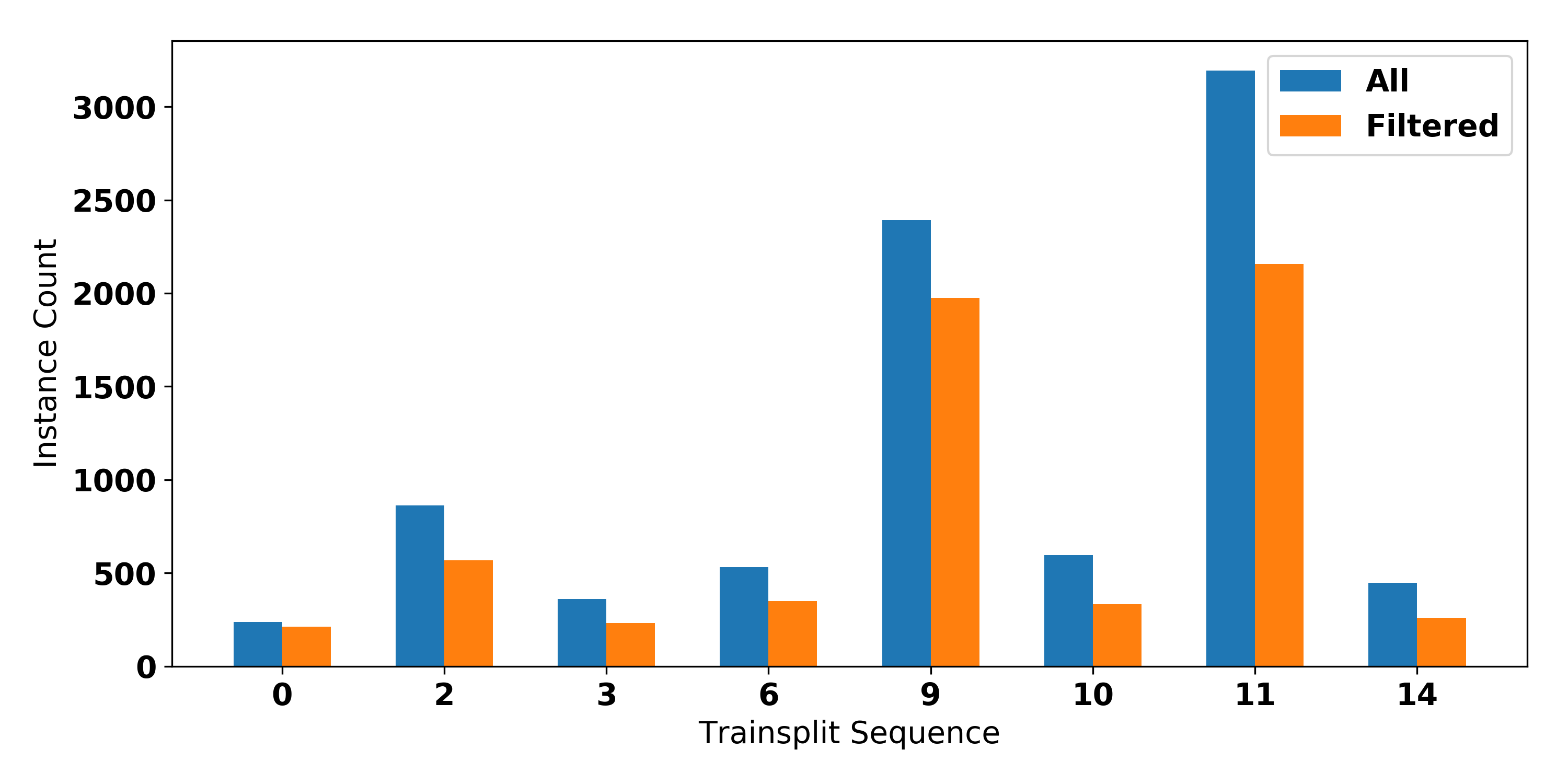}
           \caption{Percentage of the instances with accepted self-supervised depth masks, with respect to the total number of instances in each train split sequence}
      \label{fig:depth_filtered_gt}
  \end{figure}

\begin{figure*}[!htp]
\def\svgwidth{\textwidth}
\center
      \begin{minipage}[b]{0.48\textwidth}
    \includegraphics[width=\textwidth]{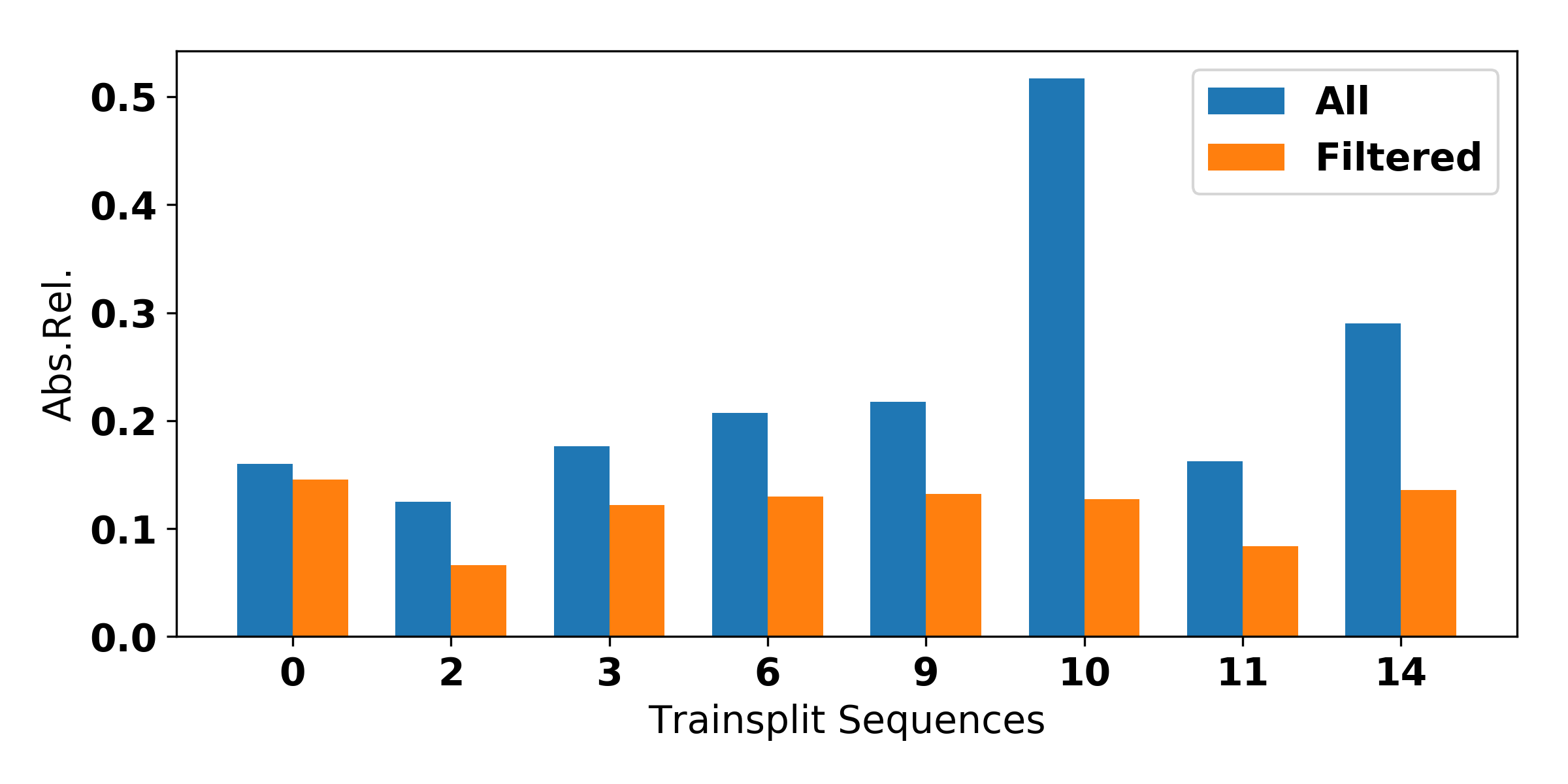}
    
  \end{minipage}
  \hfill
  \begin{minipage}[b]{0.48\textwidth}
    \includegraphics[width=\textwidth]{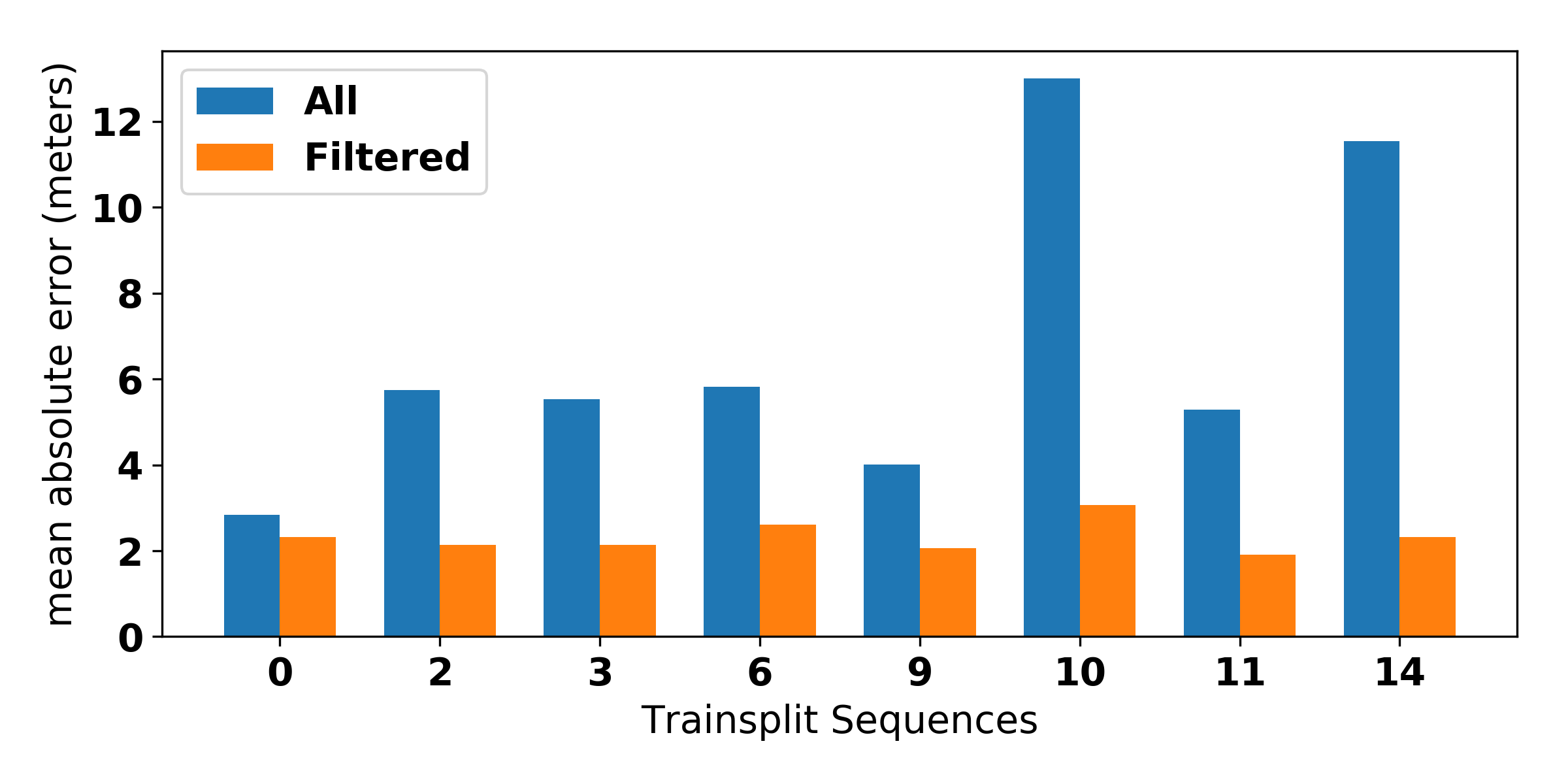}
    
  \end{minipage}

      \caption{Self-supervised depth errors as compared to the ground truth. We calculate Abs.Rel (left) or the relative absolute depth error, and the mean absolute error (right) in meters, for each instance in the ground truth. We show two bars per sequence for the filtered instances (with error less than 6 meters) and the total instances.}
      \label{fig:error_depth_gt}
  \end{figure*}

To better evaluate the cars' depth accuracy, we evaluate the center depth of the car against the KITTI train split ground truth. To this end, we first select objects by their 2D bounding box ground truth. For each object, we retrieve the mask (predicted by Mask R-CNN \cite{mask}) and use it to extract depth points from the image depth map. Specifically, in Figure \ref{fig:depth_filtered_gt}, we report the number of instances passing the filtering procedure (less than 6 meters depth difference). Additionally, in Figure \ref{fig:error_depth_gt}, we calculate the mean absolute depth error, in addition to the mean relative absolute depth error, often used in depth estimation evaluation. The last metric (typically 
 referred to as Abs.Rel) is calculated by dividing the depth error by the ground truth depth. 
Note that Figure \ref{fig:error_depth_gt} discards sequences 5 and 18 for scale issue since these sequences reach mean absolute error above $20$ meters and Abs.Rel. above $0.80$.

As noted in the paper, the pre-trained model we use achieves a mean Abs.Rel. of $0.11$ on dense depth estimation benchmark. The mean value is smaller than most means we obtain for each sequence. This is mostly due to the easier task of background depth estimation, especially road pixels whose depth is easier to learn.

\subsection{Semi-Supervised Depth Accuracy}
\begin{figure*}[!htp]
\def\svgwidth{\textwidth}
\center
      \begin{minipage}[b]{0.48\textwidth}
    \includegraphics[width=\textwidth]{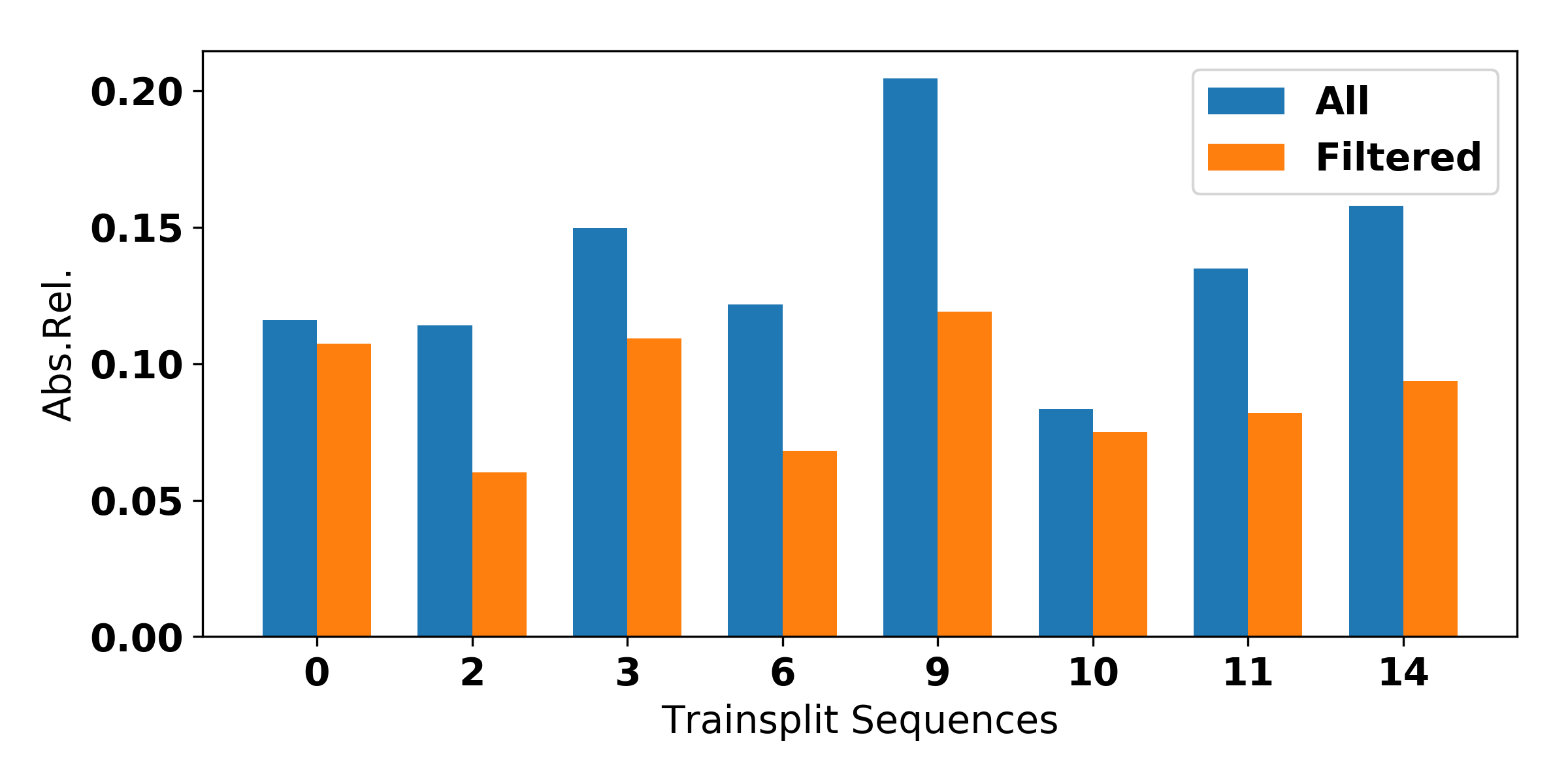}
    
  \end{minipage}
  \hfill
  \begin{minipage}[b]{0.48\textwidth}
    \includegraphics[width=\textwidth]{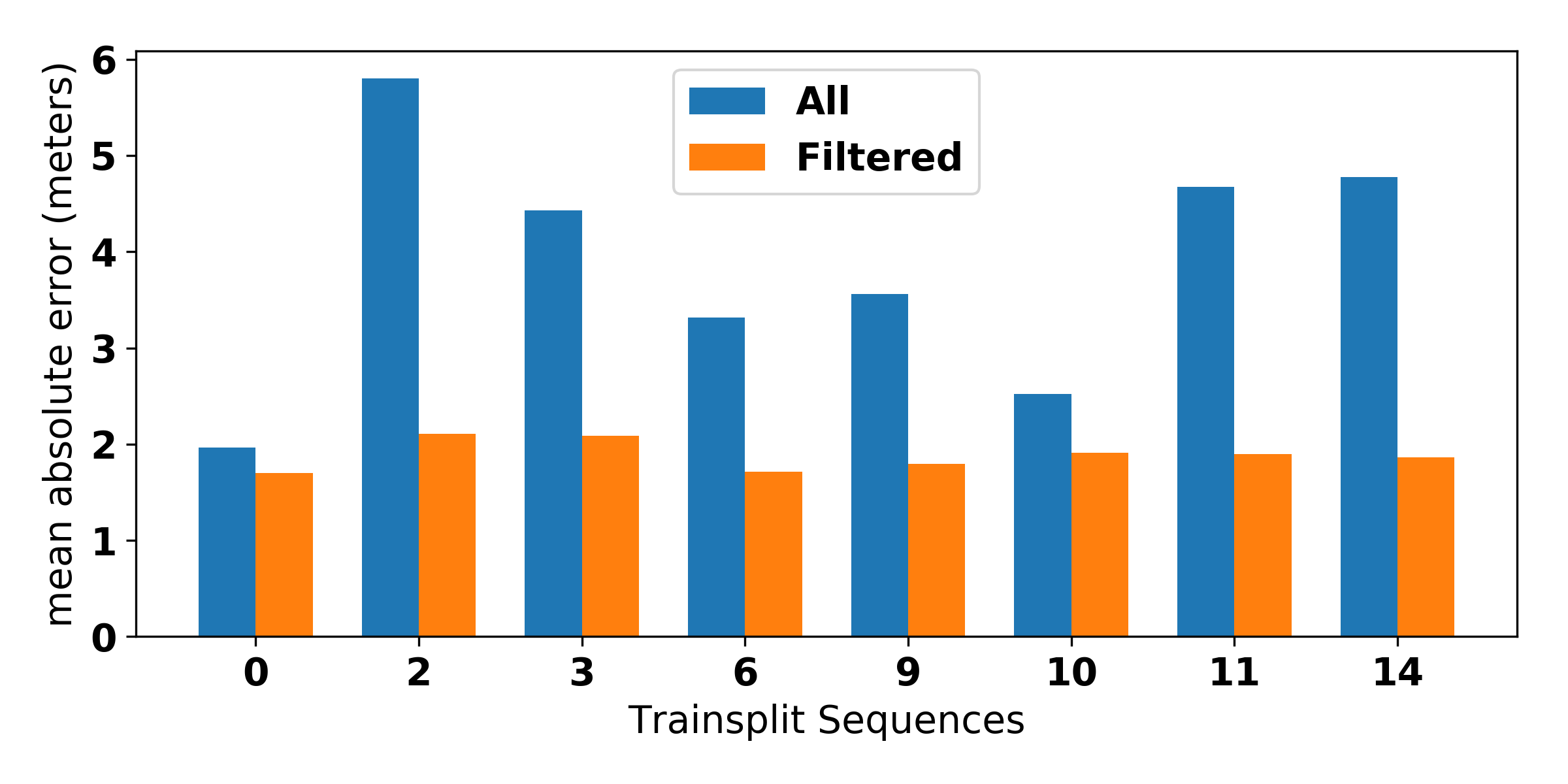}
    
  \end{minipage}
      \caption{Semi-supervised depth errors as compared to the ground truth. We calculate Abs.Rel (left) or the relative absolute depth error, and the mean absolute error (right) in meters, for each instance in the ground truth. We show two bars per sequence for the filtered instances (with errors less than 6 meters) and the total instances.}
      \label{fig:semidepth-errors}
  \end{figure*}

We repeat depth experiments from Section \ref{sec:self}, for the semi-supervised variant \cite{semidepth}. Figure \ref{fig:semidepth-errors} depicts the center depth error as computed against the ground truth, and shows a significant reduction in error values across all sequences of the train split. 
\section{Qualtitative Results}
\begin{figure*}[!htp]
\def\svgwidth{\textwidth}
\center
      \includegraphics[width=0.975\textwidth]{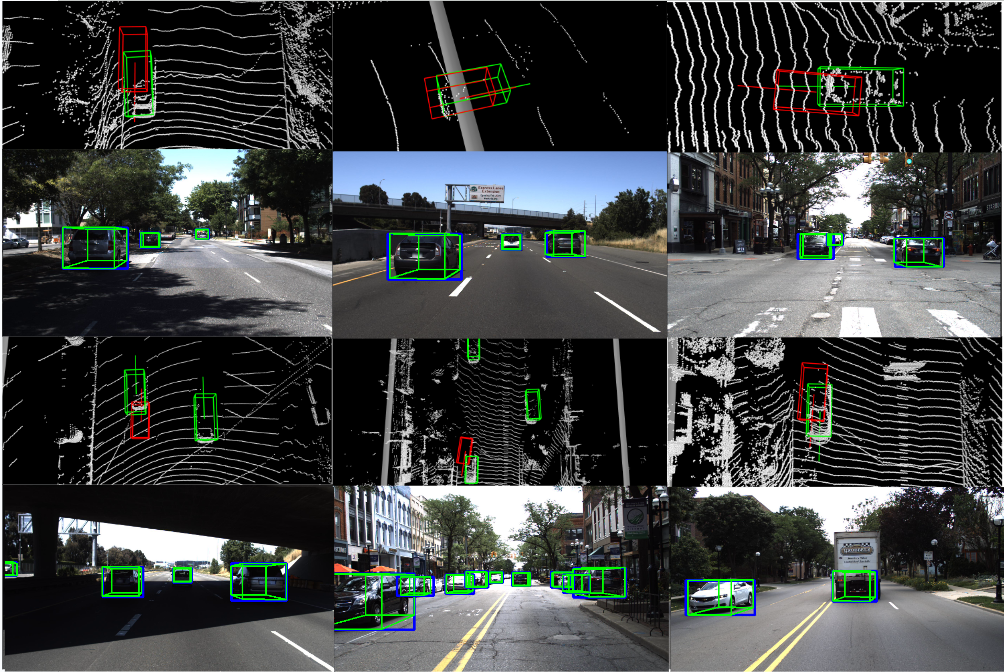}
      
      \caption{Qualitative examples of our fine-tuned model on DDAD \cite{packnet} (green) compared to the original model trained on Carla (red). 2D bounding boxes are shown on the image plane for reference (blue)}
      \label{fig:ddad_add}
  \end{figure*}

\begin{figure*}[!htp]
\def\svgwidth{\textwidth}
\center
      \includegraphics[width=0.975\textwidth]{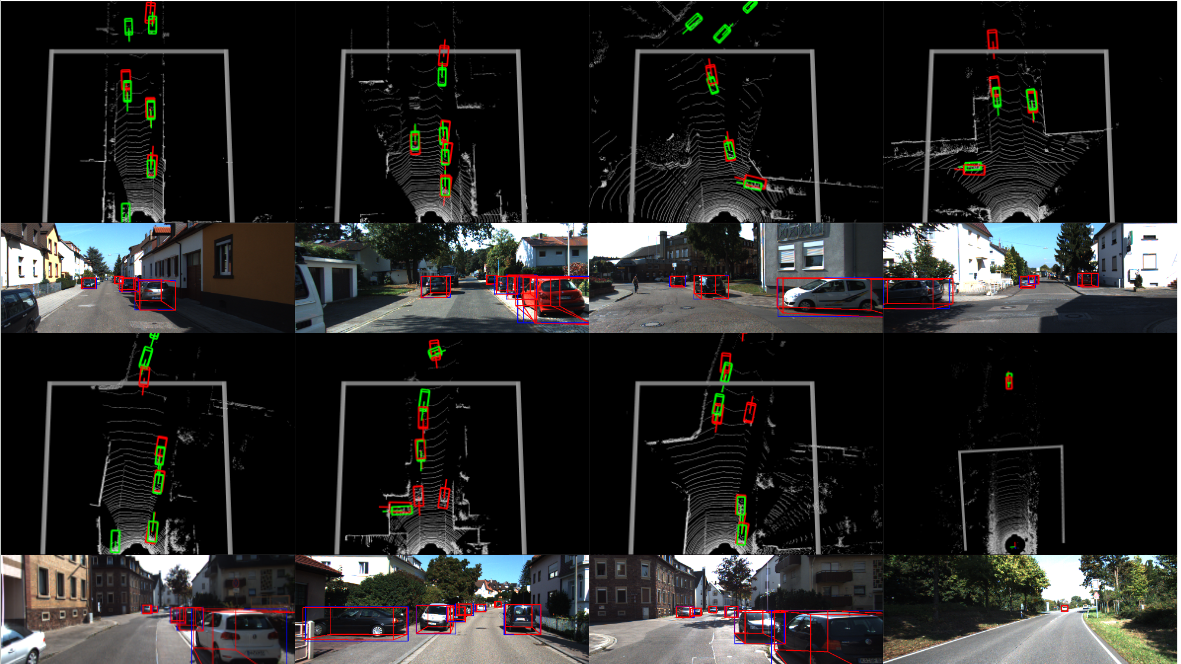}
      
      \caption{Qualitative examples of our fine-tuned model on KITTI validation set \cite{kitti} (red) compared to the ground truth (green). 2D bounding boxes are shown on the image plane for reference (blue)}
      \label{fig:kitti_qual}
  \end{figure*}

In this section, we provide additional qualitative results of the finetuned models on KITTI and DDAD. Figure \ref{fig:ddad_add} illustrates detections obtained from the finetuned model on DDAD datasets, using one iteration of pseudo-labeling, while Figure \ref{fig:kitti_qual} shows results of our best model on KITTI validation set.

{\small
\bibliographystyle{ieee_fullname}
\bibliography{egpaper_final}
}

\end{document}